\documentclass{IEEEcsmag}

\usepackage[colorlinks,urlcolor=blue,linkcolor=blue,citecolor=blue]{hyperref}

\usepackage{upmath}
\usepackage{amsfonts}

\jvol{XX}
\jnum{XX}
\paper{8}
\jmonth{July/August}
\jname{Security \& Privacy}
\pubyear{2022}

\usepackage[numbers]{natbib}
\usepackage{amsmath}
\usepackage{bbm}
\usepackage{smartdiagram}
\usepackage{environ}

\newcommand{\R}[0]{\mathbb{R}}

\begin{document}
	
	
	\title{Data Privacy and Trustworthy Machine Learning}

	\author{Martin Strobel}
	\affil{National University of Singapore}
	
	\author{Reza Shokri}
	\affil{National University of Singapore}

	\markboth{Department Head}{Paper title}
	
	\begin{abstract}
The privacy risks of machine learning models is a major concern when training them on sensitive and personal data.  Research shows that some constraints that \textit{trustworthy machine learning} poses on the training process can create new considerable privacy concerns. We discuss the tradeoffs between data privacy and the remaining goals of trustworthy machine learning (notably fairness, robustness, and explainability).
	\end{abstract}
	
	\maketitle
\chapterinitial{Machine learning algorithms} try to discover patterns from the training data that can be used to perform particular predictive tasks on new data. The application domains of machine learning touch many aspects of personal life, ranging from image processing used in health care and self-driving cars, natural language processing for personal assistants, to decision support systems in educational and judicial systems. 

In critical domains such as medicine, finance, education, and the judicial system, we need to analyze machine learning algorithms with respect to their trustworthiness, beyond their average prediction accuracy.  Can these models be trusted to operate on our personal and \textit{private} data? Are these systems \textit{fair}, or do they replicate or potentially increase existing bias in society? Can we rely on their decisions, by being able to \textit{explain} how models arrive at a decision? Can we intervene if they draw conclusions for the wrong reasons? Are the models \textit{robust} to manipulation during deployment and training, or can small changes to the data result in adversarial behavior? 

\begin{figure}
	\centering
	\includegraphics[width=0.7\columnwidth]{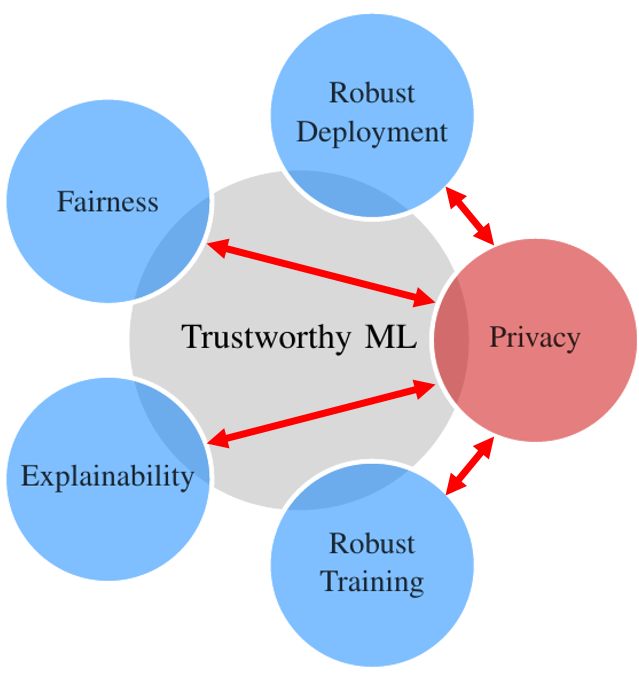}
	\caption{Different pillars of trustworthy machine learning. In this article, we discuss how data privacy interacts and conflicts with the other aspects.}
	\label{fig:trustworthy_ml}
\end{figure}

\begin{figure*}[ht!]
	\centering
	\includegraphics[width=1.8\columnwidth]{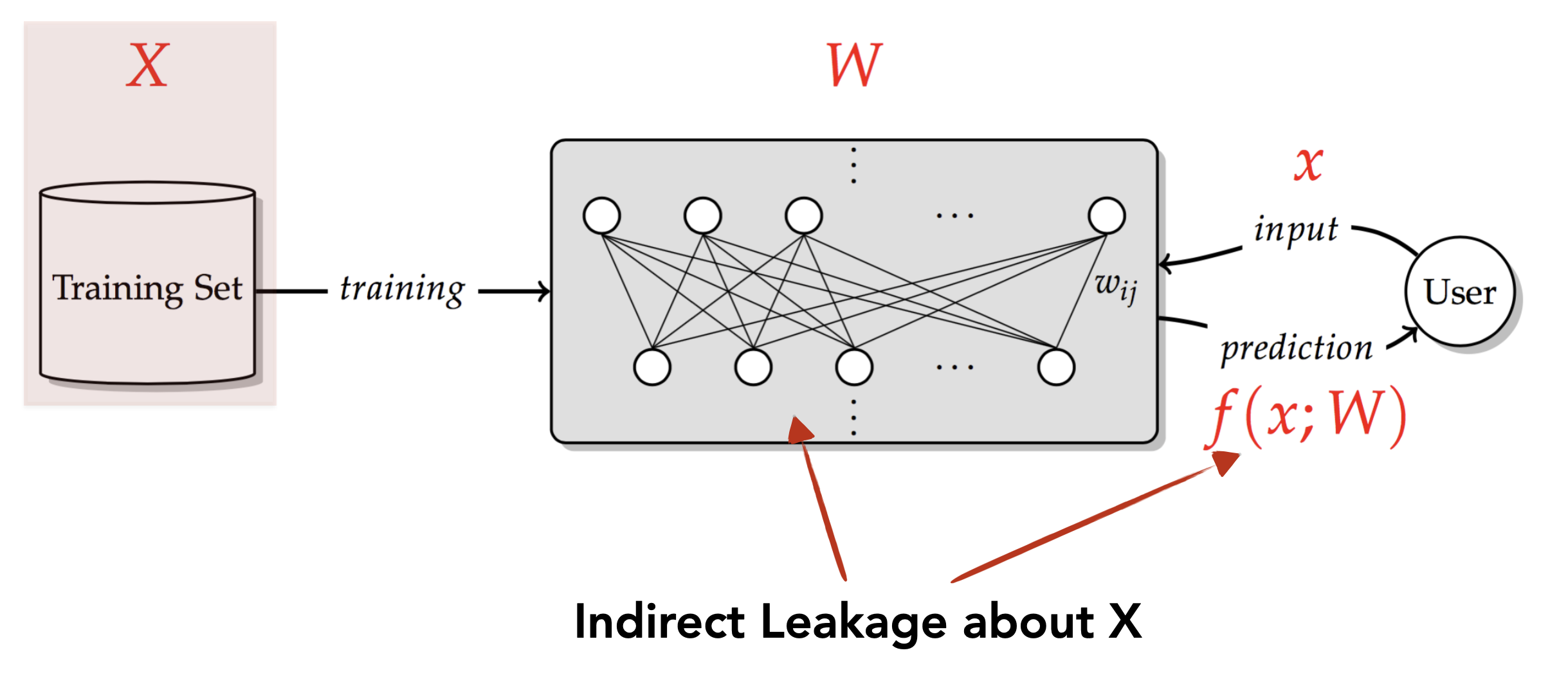}
	\caption{The typical machine learning workflow. Model parameters (learned from private training data) and model predictions can indirectly leak sensitive information about the training data. This leakage can happen even if the entire training process is confidential (e.g., using secure MPC).}
	\label{fig:privacy_threats}
\end{figure*}

To ensure \emph{trustworthy machine learning}, we need to pose additional constraints on the models we can create. We use specifically designed algorithms to make models privacy-preserving, fair, robust, or explainable. These algorithms, however, come with trade-offs. There is a significant body of research on studying the cost of trustworthy ML on the performance of models with respect to their prediction accuracy. What is very concerning is that constraining a model could also pose a risk to the privacy of training data. In this article, we discuss the interaction of data privacy with other pillars of trustworthy machine learning.

\section{Threats to Data Privacy}\label{sec:privacy}

When considering data privacy, it is important to differentiate between direct and indirect leakage of information. Direct leakage can happen during training or inference time when an untrusted entity can access the data. This threat can be prevented via access control and confidential computing (secure multi-party computation, homomorphic encryption, and trusted hardware). These measures, however, cannot prevent indirect leakage, which is the focus of this article.

An adversary, who can only observe the model and not the training data, can use inference algorithms to reconstruct information about the training data.  Obviously, it lies in the nature of useful machine learning that model parameters, and consequentially model predictions, contain \textit{some} information about the training data.

To obtain a precise definition for data privacy, we need to differentiate between general patterns that apply to the entire \textit{population}, which we would want to reveal, and the patterns that apply to specific data of \textit{individual} users, which we want to keep private. Hence, learning anything about individual data records beyond the general patterns should be considered a privacy violation.  This is the basis of \emph{differential privacy}, which measures the worst-case sensitivity of an algorithm to changes in individual users' data.

We design a (hypothetical) \emph{indistinguishably game}, between the algorithm and the adversary, to measure the amount of information leakage of an algorithm about its training data. In the game, we consider two possible worlds. In the first world, a model is trained on a dataset containing the information of a particular individual~$x$. In the second world, the model is trained on the same dataset, except this time the ~$x$'s data is removed. The adversary interacts with the model, in an unknown world, and tries to infer if the model's training set contains~$x$'s data. The capacity to which the adversary can win this game tells us the extent of individual private information leakage. In the white-box setting, the attacker has complete access to the model's parameters. In the black-box setting, the attacker can only interact with the model's predictions.  \citet{shokri2017membership} designed \emph{membership inference attacks}, which are algorithms to simulate the game and measure the information leakage of models about their training data. Besides, the reports by the US National Institute for Standards and Technology (NIST) \cite{nist2020taxonomy} and the UK Information Commissioner's Office (ICO) \cite{ioc2020guidance} specifically mention membership inference as a confidentiality violation
and potential threat to the training data in AI.  

In the black-box setting, membership inference attacks attempt to exploit the signals contained in a model's predictions. Major examples of such a signal are the prediction error and uncertainty. Predictions tend to be more accurate and certain for members of the training set compared to unseen points.  Powerful membership inference attacks have been demonstrated for many different scenarios in machine learning~\cite{nasr2018comprehensive, ye2021enhanced}.

We can use membership inference attacks to quantitatively measure the privacy risks of machine learning algorithms in many diverse scenarios. Throughout this article, we will use membership inference attacks as a tool to measure information leakage of private data. This way, we can study the privacy implications of the other aspects of trustworthy machine learning.

Given a set of points $X \subseteq \R^n$ with membership $m\colon X \to \{0,1\}$ to a training set, we define the average \emph{privacy risk} of a trained model as\footnote{See \citet{ye2021enhanced} for a more comprehensive discussion on how to measure privacy risk in machine learning.} 
\begin{align*}
	\max_{A \in \mathcal{A}} \frac{1}{|X|} \sum_{x \in X} \left[A(s(x)) = m(x) \right].
\end{align*}

Here, $s\colon X \to \R^k$ is the signal the attacker can observe for each point after a model is trained. The signal can be the model's prediction, the model's loss, an explanation, or similar observable signals about a data point (which differentiates members from non-members). $\mathcal{A}$ is the set of possible attack algorithms the attacker can use. For actual evaluations \cite{shokri2017membership, shokri2021privacy, chang2021privacy, song2019privacy, sablayrolles2019white} specific attack algorithms are designed. Usually, $X$ is assumed to be balanced, i.e., the lower bound on the privacy risk is $0.5$ obtained by an attacker randomly guessing membership. 

A common attack strategy are \emph{threshold-based attacks} \cite{sablayrolles2019white, ye2021enhanced}. Here, the attacker directly compares the signal to a predefined threshold $\tau$. For example, it is common for the loss of training set members to be lower than for non-members. In a threshold-based attack using the point's loss as a signal, each point with a loss lower than the threshold is considered a member, and each point with a loss higher is considered a non-member. The attacker can obtain the threshold by observing the loss distribution of points on reference (or shadow) models. All experimental results presented here were originally obtained using threshold-based attacks.  

\section{Algorithmic Fairness}

\begin{figure}
	\centering
	\includegraphics[]{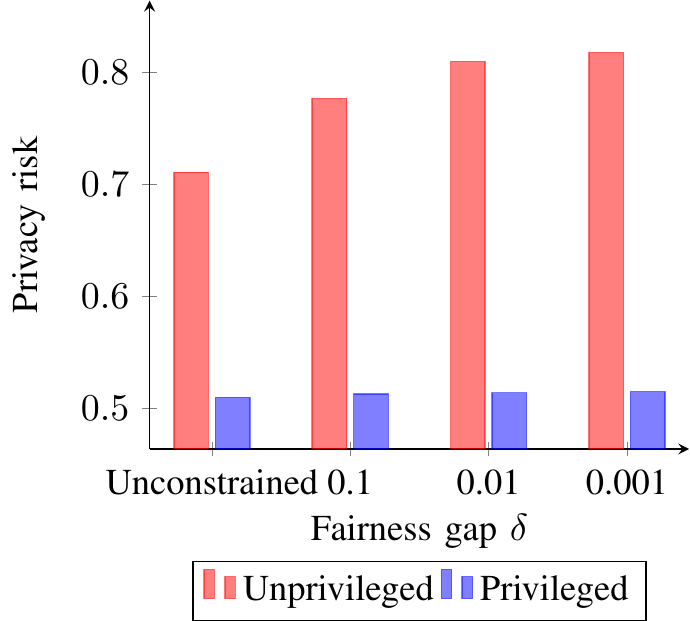}
	\caption{Experimental results on a Law School Admission problem provided by \citet{chang2021privacy}. There is a tradeoff between privacy and fairness of a model. Increasing fairness (i.e., decreasing $\delta$) increases the privacy risk of the qualified unprivileged group. }
	\label{fig:fairness_tradeoff}
\end{figure}

Machine learning models potentially cause disproportionate harm to specific \textit{groups}. This bias might arise from demographic disparities in the training data, the implicit focus on optimizing performance on the majority group, or at other steps in the machine learning pipeline~\cite{barocas-hardt-narayanan}.

While there are many different fairness concerns, the research community has paid the most attention to statistical notions of \textit{group unfairness} for classification tasks. For this, we observe the differences in the behavior of a classifier for inputs of different demographic groups according to specific protected attributes (e.g., gender, race). In this context, protected attributes are provided by anti-discrimination laws or societal standards. 

As an illustrative example, let us consider the following hypothetical scenario: A company wants to use an automated classifier to approve loan applications based on various user-provided data. An auditor can split a dataset for which the classifier has made approval decisions into the decisions made for male and female applicants. One criterion of fairness would be to require that applicants that pay back their loan have an equal opportunity of getting the loan in the first place. This notion of fairness is known as \emph{equal opportunity}~\cite{hardt2016equality}. Formally, we can say that a classifier $f\colon \R^n \to \{0,1\}$ satisfies $\delta$-Equal Opportunity with respect to two groups $X_1,X_2$ and true labels $y\colon \R^n \to \{0,1\}$, if the false negative rate of the classifier in the groups are within $\delta$ range of one another:
\begin{align*}
	&|\mathbb{P}_{x \sim X_1}\left[ f(x) = 0 | y(x) = 1 \right]\\  &-  \mathbb{P}_{x \sim X_2}\left[ f(x) = 0 | y(x) = 1 \right]  | \leq \delta.
\end{align*}

Note that equal opportunity is only one of many potential fairness notions. The auditor could also compare false-positive rates among the groups, or the auditor could validate if the approval rates are the same among the genders. 

Fairness-aware learning is a way to mitigate the bias of a given classifier, by enforcing the fairness constraints. These algorithms can either modify the training set and the training algorithm, or post-process the fully trained model.  \\\

\textit{\textbf{``Achieving group fairness comes at a cost for individual privacy.''}}\\

\citet{chang2021privacy} show that the achieved group fairness comes at a cost for privacy. They create a group-specific membership inference attack to demonstrate that the privacy risks of different demographic groups are affected disparately by fairness-aware learning. In particular, the privacy risks of the unprivileged group (i.e., the group the original classifier is biased against) increases. Higher bias in the unconstrained model leads to greater privacy risks of the unprivileged class in the corresponding fair model. Further, the stronger the fairness constraint is enforced, the more the privacy risks increase. 

Figure~\ref{fig:fairness_tradeoff} reproduces their experimental results on how enforcing a smaller fairness gap increases the privacy risk. The experiment considers a depth-10 decision tree trained on the law school admission dataset. Fairness is enforced using the reductions approach (see \cite{chang2021privacy} for more details). 

Why is there this tradeoff between privacy and fairness? An answer lies in how we create unbiased models. By enforcing constraints, fairness-aware algorithms ensure the equal performance of the model on different subgroups. However, correctly learning the underprivileged group's classification might be difficult for several reasons. There might be fewer data available for the group, the data might have higher variance, or the task might be more complex. Hence, this might force the model to \textit{memorize} more of the underprivileged group's data instead of correctly learning a general pattern. This higher memorization links directly to higher privacy risk. It is easier for the adversary to differentiate between members and non-members. It follows that group fair models have a higher privacy risk.

\citet{bagdasaryan2019differential} provide complementary evidence for the privacy-fairness tradeoff. They demonstrate that the accuracy reduction caused by differentially private training disparately affects different groups. Smaller, unprivileged subgroups suffer more. Again the intuitive connection stems from memorization. Smaller groups rely more on memorization, which differentially private training suppresses. 

For some settings, this tradeoff is even theoretically unavoidable. \citet{cummings2019compatibility} show cases where a model cannot be trained in a differentially private way and still satisfy group fairness constraints exactly.

Overall we have strong experimental \cite{bagdasaryan2019differential, chang2021privacy}  and theoretical \cite{cummings2019compatibility} evidence on the incompatibility of individual privacy and group fairness.

\section{Explainability}
Modern machine learning often achieves performance improvements by increasing the complexity of the model architectures and training process. In consequence, the patterns learned by the models are harder to understand. Further, it becomes difficult or impossible to comprehend why a model came up with a specific decision. However, this lack of understanding is undesirable for many critical decision-making scenarios. Here, we require that the semantics of the model align with the semantics of the tasks. This way, we can be confident that a model did not pick up spurious correlations in the training data \cite{molnar2020interpretable}.

\emph{Feature-based} attribution methods are model explanations that try to highlight which features of the input were important to a specific point's prediction. Many of these attribution methods are based on the model's gradient with respect to the input. The gradient tells us exactly how much a model's output changes if we make infinitesimally small changes to input. So methods derived from the gradient highlight influential input features.\\

\textit{\textbf{``Model explanations can be exploited by inference attacks.''}}\\

From the privacy perspective, model explanations provide the attacker with an additional source of information. This is especially true in scenarios where the attacker cannot directly access a model's uncertainty or loss. High feature attribution values are a proxy for model uncertainty. They indicate that a small change to the input would radically change the model's output. Hence, an attacker can construct successful membership inference attacks solely based on model explanations to distinguish between members and non-members~\cite{shokri2021privacy}.

\begin{figure}
	\centering
	\includegraphics[]{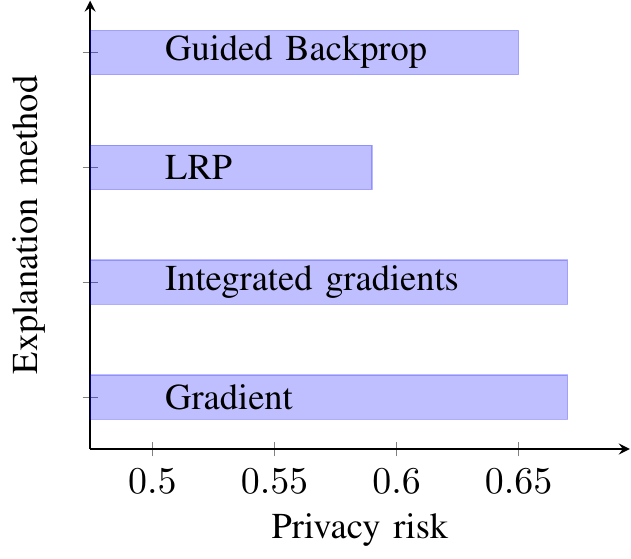}
	\caption{The privacy risks of different backpropagation-based explanation methods of a neural network trained on the Texas hospital dataset. Provided by \citet{shokri2021privacy}.}
	\label{fig:explanation_attacks}
\end{figure}

Figure~\ref{fig:explanation_attacks} shows the privacy risks for a membership inference attack based on different explanations methods for a neural network trained on the Texas hospital dataset (see \cite{shokri2021privacy} for more details). So, model explanations can also conflict with individual privacy.

\section{Robustness during deployment}

\begin{figure}
	\centering
	\includegraphics[]{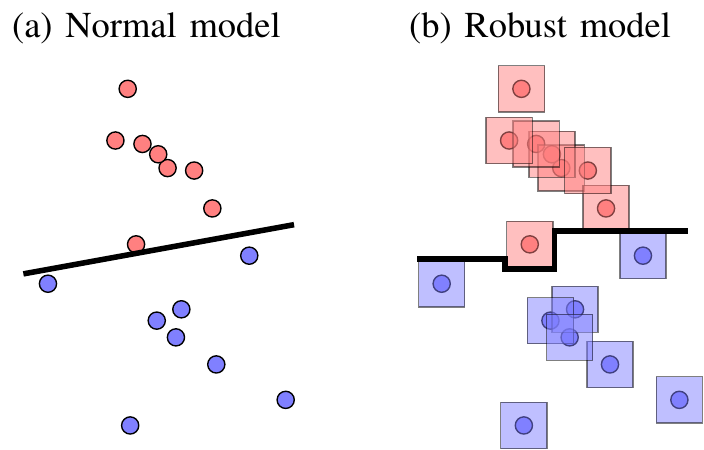}
	\caption{Many of the existing robust training algorithms change the decision boundary around training points in a specific way. It is possible for an attacker to discover and exploit these patterns to create more successful inference attacks.}
	\label{fig:robust_models}
\end{figure}

Decision boundaries of modern neural networks are very complex. For many normal inputs, an adversary can craft a very small change in a specific direction. The manipulation is almost unnoticeable yet still crosses the model's decision boundary. So, adversaries can force wrong classifications and potentially bypass security checks. 

So, a robust model is required not just to correctly predict the label for normal and benign inputs, but to remain correct if the input is slightly manipulated. For a distribution $(x,y) \sim \mathcal{D}$ of points $x$ and labels $y$, the \emph{standard supervised learning objective} is to find a parameter $\theta$ that minimizes the expected loss $L$ of a classifier $f_\theta$
\[
\min_\theta \mathbb{E}_{(x,y) \sim \mathcal{D}} \left[  L(f_\theta(x,y) \right].
\]
Assuming that an adversary can modify each point by $\delta \in S$ the \emph{robust supervised learning objective} is to minimize the maximal loss an attacker can achieve \cite{madry2017towards}:

\[
\min_\theta \mathbb{E}_{(x,y) \sim \mathcal{D}} \left[ \max_{\delta \in S} L(f_\theta(x+\delta,y)) \right] 
\]

Many defenses against evasion attacks affect the model's decision boundaries so that, within a small area around each input, model predictions remain the same~\cite{madry2017towards}. Yet, the methods can enforce this objective only on the training data. It follows that the impact of individual training points on robust models increases. The more extensive influence raises the privacy risk of training set members.\\

\textit{\textbf{``Defending against evasion attacks (adversarial examples) via adversarial training can increase privacy risks.''}}\\

\begin{figure}
	\centering
	\includegraphics[]{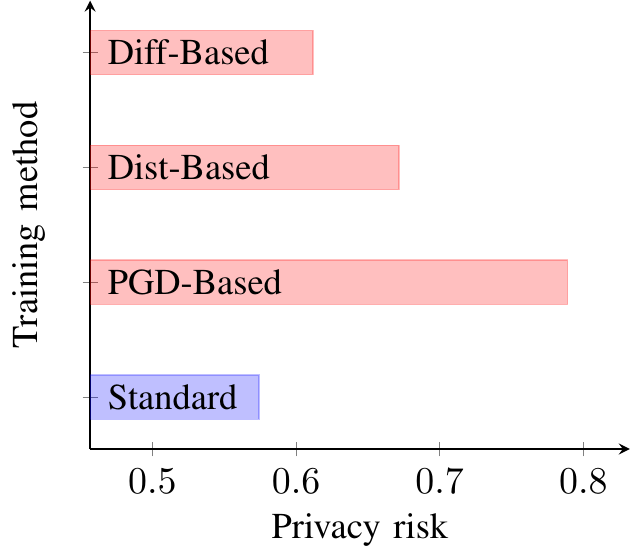}
	\caption{The privacy risks of robust-training methods compared to standard training of a neural network trained on the CIFAR10 dataset. Provided by \citet{song2019privacy}.}
	\label{fig:robustness_attacks}
\end{figure}

In Figure~\ref{fig:robustness_attacks}, we reproduced \citet{song2019privacy}'s results for the privacy risks of different robust training methods for a neural network trained on the CIFAR10 dataset. The entire current generation of robustness algorithms that change the model's training behavior seems to suffer from this robustness-privacy tradeoff. However, it is unclear whether there exists a fundamental tradeoff between robustness to attacks and privacy. 

\section{Robustness during training}

The quality and reliability of the training data cannot be guaranteed in all scenarios. Crowdsourced data might suffer from unmotivated or malicious crowd workers. Data collection might be noisy. In a federated setting, data from different clients is not of the same quality. In the worst case, an adversary might actively try corrupting parts of the training data to manipulate the training process and harm the model's performance.

Among all threats \emph{data poisoning} attacks have gotten the most attention. We can only consider the model's training robust if it has some form of immunity to data poisoning attacks. Modern machine learning algorithms are generally not robust against such threats. Many works have demonstrated the possibility of manipulating models via poisoning their training data. 

Some robust training algorithms try to reduce the sensitivity of the training to changes of small parts of the training set. Others try to detect and completely ignore outliers. The goal is to make the model's training focus on general patterns and avoid being influenced by distinct points.\\

\textit{\textbf{``Robustness against poisoning attacks aligns well with data privacy.''}}\\

In contrast to the other goals of trustworthy machine learning, the constraints posed by training data robustness align well with data privacy. Both differential privacy (DP) and robustness are notions of insensitivity to training data changes. While the existing metrics for these notions are not exactly the same, DP focuses on worst-case changes of single inputs and robustness on restricted changes of groups of data. There are reasons for cautious optimism. Research of robust statistics further supports this optimism as robust mechanisms are demonstrably good starting points for differentially private mechanisms \cite{dwork2009differential}. 

\section{Conclusions}

In this article, we have discussed many aspects in which current ways to ensure a more trustworthy classifier come at a cost to data privacy. Yet, the situation is not clear-cut. While group fairness comes with a fundamental tradeoff to privacy~\cite{cummings2019compatibility, chang2021privacy}, for the other aspects, much is unknown. Current algorithms to make models robust against attacks during deployment hurt data privacy \cite{song2019privacy}. But we cannot rule out the possibility of designing robust algorithms which are also accurate and privacy-preserving. The situation is similar in the case of explainability. Here we also understand that some existing methods are vulnerable \cite{shokri2021privacy}. Yet, we do not know if it extends to all types of explanations.

Trustworthy machine learning is a laudable goal, but it should not come at the cost of data privacy. We call for more investigation of the interactions of these two aspects. Instead of looking at these problems separately, we need to develop techniques to achieve both.  

Finally, privacy concerns in machine learning are not limited to membership inference. Researchers have considered further threats such as reconstruction attacks, model extraction, or property inference. These concerns also interact with the other aspects of trustworthy machine learning and require further study.

\begin{IEEEbiography}{Martin Strobel}{\,} is a Ph.D. candidate at the National University of Singapore's School of Computing working under the supervision of Yair Zick and Reza Shokri. Before that, he obtained his Master of Mathematics, a Bachelor's degree in Computer Science, and a Bachelor's degree in Mathematics from the Technical University of Munich.	His research interests are in the privacy and transparency aspects of machine learning.
	
\end{IEEEbiography}

\begin{IEEEbiography}{Reza Shokri,}{\,} is a NUS Presidential Young Professor of Computer Science. His research focuses on data privacy and trustworthy machine learning. He is a recipient of the IEEE Security and Privacy (S\&P) Test-of-Time Award 2021, for his paper on quantifying location privacy. He received the Caspar Bowden Award for Outstanding Research in Privacy Enhancing Technologies in 2018, for his work on analyzing the privacy risks of machine learning models. He received the NUS Early Career Research Award 2019, VMWare Early Career Faculty Award 2021, Intel Faculty Research Award (Private AI Collaborative Research Institute) 2021, and Meta Faculty Award 2021. He obtained his PhD from EPFL.  
\end{IEEEbiography}

\end{document}